\documentclass[runningheads]{llncs}
\usepackage{graphicx}
\usepackage{xcolor}
\begin{document}
\title{Evaluation of Neural Network Classification Systems on Document Stream}

\author{Joris Voerman\inst{1,2} \and
Aur\'elie Joseph\inst{2} \and
Mickael Coustaty\inst{1} \and
Vincent Poulain d'Andecy\inst{2} \and
Jean-Marc Ogier\inst{1}}

\authorrunning{J. Voerman et al.}

\institute{La Rochelle Universit\'e, L3i \\
  Avenue Michel Cr\'epeau, 17042 La Rochelle, France\\
  \email{\{firstname.lastname\}@univ-lr.fr} \and
Yooz \\
  1 Rue Fleming, 17000 La Rochelle, France\\
  \email{\{firstname.lastname\}@getyooz.com}}

\maketitle

\begin{abstract}
One major drawback of state of the art Neural Networks (NN)-based approaches for document classification purposes is the large number of training samples required to obtain an efficient classification. The minimum required number is around one thousand annotated documents for each class. In many cases it is very difficult, if not impossible, to gather this number of samples in real industrial processes. In this paper, we analyse the efficiency of NN-based document classification systems in a sub-optimal training case, based on the situation of a company’s document stream. We evaluated three different approaches, one based on image content and two on textual content. The evaluation was divided into four parts: a reference case, to assess the performance of the system in the lab; two cases that each simulate a specific difficulty linked to document stream processing; and a realistic case that combined all of these difficulties. The realistic case highlighted the fact that there is a significant drop in the efficiency of NN-Based document classification systems. Although they remain efficient for well represented classes (with an over-fitting of the system for those classes), it is impossible for them to handle appropriately less well represented classes. NN-Based document classification systems need to be adapted to resolve these two problems before they can be considered for use in a company’s document stream.
\end{abstract}

\section{Introduction}
\label{sec:introduction}
Companies generate a large amount of documents everyday, internally or external entities. Processing all of these documents required a lot of resources. For this reason, many companies use automatic system like Digital Mailroom\cite{INTELLIX} to reduce the workload of those documents processing. Companies like ABBYY, KOFAX, PARASCRIPT, YOOZ, etc. propose performing solutions that can process electronic and paper documents. For the classification task, most of them use a combination of several classifiers specialized for one or some type of documents, with at least one based on Machine Learning techniques. However, all these systems have the same issue: they need to be retrained each time new classes are added. This maintenance is costly in time and in workload because it needs to build a learning dataset. An option to overcome this problem is to use an incremental learning system like \cite{INTELLIX} or \cite{INDUS}, but they are not able to compete with state of the art deep learning systems in performances.

One main objective of this paper is to explore the adaptation of neural network systems to Digital Mailroom context for classification and extraction of main information inside all companies documents. Thereby, in such industrial context, an error (i.e. a misclassified document) has more impact than a rejection (i.e. a document rejected by the system and tagged with no class) because when a document is rejected, an operator will be warned and will correct it. On the opposite, an error will not be highlighted as a rejection and the error will be propagated into the next steps of the system.

One of the best possible modelization of Digital Mailroom entries is a document stream model. A document stream is a sequence of documents that appear irregularly in time. It can be very heterogeneous and composed of numerous classes unequally represented inside the stream. Indeed a document stream is generally composed, in a first time, of a core group of some classes highly represented, that is the majority of the stream. In a second time, remaining documents are unequally distributed between the majority of other classes, less represented than previous. Many classes are composed of only few documents that do not allow an efficient training. In addition, the class composition of a document stream is in constant growing, spontaneous new category or new sub-category variation appear time-to-time. And finally, the number of documents per class increases as the content of a document stream evolves.

This definition highlights two main constraints. First, in a real document stream application, classes representation are unbalanced. When a training set is generated from a document stream, it will be by nature unbalanced. This could reduce neural network methods performances if low represented classes are insufficiently trained and impact higher represented classes performances like noises. Then, no train set generated from a document stream could represent all real cases. The training phase will then be incomplete because the domain changes endlessly. These unexpected/incomplete classes, at first glance, cannot be managed by a neural network system.

The objectives of this experimentation is to evaluate the adaptation of neural network classification methods to these two constraints. Also, to determine and quantify impacts on network training phases and possible modifications that can be applied on neural network systems to counter or lessen this impact.

The next section will overview related works and methods that will be compared in the experimentation from section 4. Then, we will describe the testing protocol used by this evaluation in the section 3 and we will conclude and open perspectives in the section 5.

\section{Related Work}
\label{sec:related-work}
\subsection{Overview}
Solutions for document classification by machine learning methods can be divided in two approaches: Static Learning and Incremental Learning. Static Learning is based on a stable training corpus, supposed representative of the domain. Mostly, neural network methods follows this approach and can be themselves distinguished in two categories, regarding the fact they are whether based on pixels information or texts.

The first step to process texts with a neural network is the word representation. The most widely used techniques currently rely on the word embedding approach, like Word2Vec\cite{W2V}, GloVe\cite{GLOVE} and more recently BERT\cite{BERT}. These methods enhance previous textual feature like bag-of-words and word occurrences matrix. To limit potential noises, a word selection is commonly apply with basically a stop-word suppression ("the","and", etc.). More advanced strategies used information gain, mutual information\cite{M&GInfo}, L1 Regularization\cite{L1} or TF-IDF\cite{TF-IDF} to select useful feature. The second step is the classification itself with multiple model, mainly Recurrent Neural Network (RNN) \cite{RNN} and Convolutional Neural Network (CNN) \cite{TCNN1}\cite{TCNN2}, and more recently a combination of these two structures, named RCNN, like in \cite{RCNN}. The RNN approaches are generally reinforced by specific recurrent cell architecture like Long Short-Term Memory LSTM\cite{LSTM} and Gated Recurrent Unit GRU\cite{GRU}. Some industrial application use this type of network like CloudScan\cite{CloudScan} for invoice classification and information retrieval.\\

For the image classification, the principal category of neural network method used is currently the pixel-based CNN with a high diversity of structures: ResNets\cite{ResNet}, InceptionResNet\cite{Inception}, DenseNets\cite{DenseNet}, etc. But these approaches are not only restrictive to image and can be extended to documents classification without any text interpretation, as the RVL-CDIP dataset challenge\cite{HCNN}.

All these methods are highly efficient for document classification when the Static Learning condition is met, but as explained in the introduction, document stream does not encounter this condition. The second approach has been designed to solve this issue with an Incremental Learning process. The main idea is to reinforce the system gradually at each data encounter and no more in one training phase. The incremental approach for neural network is relatively recent because their current structure were not designed for this task. However, two potential approaches have emerged recently: Structural Adaptation\cite{StrucAdapt} or Training Adaptation\cite{TrainAdapt1}\cite{TrainAdapt2}. Out of neural network, several classifiers based on classical algorithm can be find like incremental SVM\cite{IncSVM}, K-means\cite{IncKM} and Growing Neural Gas (IGNG)\cite{A2ING} or industrial applications like INTELLIX\cite{INTELLIX} or InDUS\cite{INDUS}.\\

In addition of previous methods, two trends appeared in the last decade and are closed to our situation for classification of low represented classes: Zero-shot learning and One-Shot/Few-Shot Learning.

The first, Zero-Shot learning is an image processing challenge where some classes are need to be classified without previous training samples\cite{ZSOver}. This description seems to be perfect for document stream classification, but all methods are based on transfer learning from a complete domain, mainly textual, where all classes are represented. This cannot be applied because no complete domain exists for our case.

The second, One-Shot Learning, is also an image processing challenge but where some classes are represented by only few, at least one, samples. This is the case for some low represented class from document stream. Two main groups of approaches exist: first group relies on the use of Bayesian-based techniques like in \cite{OS1} and \cite{OS2} ; the second group relies on modified neural network architecture like the Neural Turing Machine (NTM), the Memory Augmented Neural Network (MANN)\cite{NTM} and the Siamese Network\cite{SiamNet}.

In order to assess the performances of those categories of approaches on the dedicated case of document stream classification, we propose in the next section to provide a more detailed description of the evaluated methods. These ones have been chosen for their specificities that will be presented hereafter.

\subsection{Compared Methods}
\label{sec:method}

\subsubsection{Active Adaptive Incremental Neural Gas (A2ING) \cite{A2ING}}.\\
A2ING is a document stream classification method by Neural Gas that use an active semi-supervised sequential training.

The classification by Neural Gas is a machine learning method inspired by human brain. All classes are represented by centroids, called neurons. These centroids are put in a space composed in accordance with features chosen to describe the data, here documents. A document is then associated to the closest centroid in this feature space. The training is done on the position of centroids in the space and the classification range. This range is associate to each centroid and limit the distance allowed to express themselves. This range limitation is equally used to detect new classes.

This version is inspired by the Adaptive Incremental Neural Gas (AING) method, which was designed to relies on an incremental learning with an adaptive training phases. A2ING is a version with a active semi-supervised sequential training. Active mean that the training phases need an operator to correct algorithm answers because the learning dataset contains labelled and unlabelled data, called semi-supervised.

INDUS\cite{INDUS}, developed by the french company YOOZ, is design to classify document stream. This system relies on the A2ING classification algorithm, and uses the textual information as feature to characterize documents. More precisely, it used the TF-IDF. INDUS has also a souvenir system allow remembering classes with few documents in order to assess performance improvement during the training phase. This study compare it with neural network on document stream classification.

\subsubsection{Holistic Convolutional Neural Network (HCNN) \cite{HCNN}}.\\
HCNN is an image classifier using a CNN based on the pixels information. Its weights are initialized by a model trained on ImageNet 2012 challenge database\cite{ImageNet}. According to authors, this fine-tuned model offers better performances and this method has the best results of their study on the RVL-CDIP dataset proposed in the same article. This method take as input a fixed sized image, resized if necessary, and train pixel-level feature by multiple CNN layers to distinguish classes. The classification itself is done by three successive fully-connected layers. 
We chose the HCNN approach as a baseline for RVL-CDIP classification task using only the image information.

\subsubsection{Recurrent Convolutional Neural Network (RCNN) \cite{RCNN}}
\begin{figure}
\includegraphics[width=\textwidth]{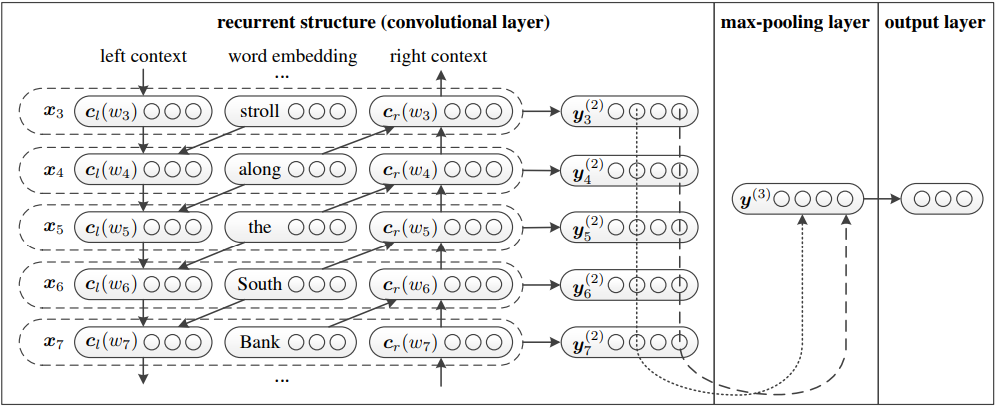}
\caption{The structure of the recurrent convolutional network scheme by \cite{RCNN} with sentence sample.} \label{fig1}
\end{figure}

.\\
RCNN is a bidirectional recurrent CNN for text classification. It uses a word embedding like word2vec skip-gram model for text representation and is divided in two steps. The first step is a bidirectional RNN, our implementation used LSTM cell in the RNN layers. This network compute completely the two sides context of each word to enhance the word embedding. Contexts are the right and left sides of the word and are computed from the beginning to the end of the text. The second step is a CNN feed by the result of the bi-RNN(\textbf{Fig.\ref{fig1}}). This network end in a fully connected layer to classify input text.

\subsubsection{Textual Convolutional Neural Network (TCNN) \cite{TCNN1}\cite{TCNN2}}.\\
This system is a combination inspired by two CNN designed for text classification. The combination relies on a character-level CNN\cite{TCNN2} where each layer is replaced by a multi-channel CNN\cite{TCNN1}. The result in a strong text classifier by vocabulary that has more basics feature than RCNN. This could have an importance in this evaluation with a non-optimal training set.



\section{Testing Protocol}
\label{sec:testing-protocol}
\subsection{Dataset}
The data-set used for this evaluation is RVL-CDIP \cite{HCNN}, which is a subset of the IIT-CDIP \cite{IIT-CDIP} data-set. It is composed of 400 000 documents equally distributed in 16 classes. Classes correspond to a variety of industrial and administrative documents from the tobacco’s companies. More specifically, the 16 classes are: letter, memo, email, file-folder, form, handwritten, invoice, advertisement, budget, news, article, presentation, scientific publication, questionnaire, resume, scientific report and specification. Some classes does not contain any usable text, like the file-folder one, or very few text like presentation. On the contrary, the scientific reports are mainly composed of text. Moreover, this advertisement class have an high structural diversity unlike the resume. For each class, 20 000 documents as used for the training set, 2500 for the validation set and 2500 for the test set. Originally, images have variable sizes, a resolution of 200 or 300 dpi and could be composed of one or multiple pages. For this work, we standardized all images to 754*1000 pixels, 72 dpi and one page.

In order to evaluate the language processing based methods, we applied a recent OCR software on the IIT-CDIP equivalent images for a better quality. The creator of RVL-CDIP data-set originally recommends to use their IIT-CDIP OCR text file but they were not organized in the same way than the RVL-CDIP images. Text files are computed on multiple page documents that was not in RVL-CDIP with an old OCR (2006). The new text files solve all those problems as they are computed on only one page with ABBYY-FineReader (version 14.0.107.232).

\begin{figure}
\includegraphics[width=\textwidth]{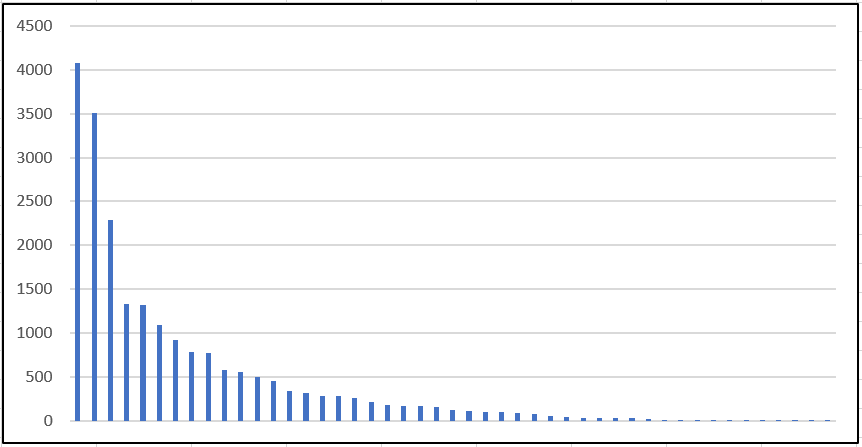}
\caption{Documents distribution between classes in Yooz private dataset. This introduce the number of document associated to each of 47 dataset classes} \label{fig2}
\end{figure}

In addition, a private dataset provided by the Yooz company was used to challenge its own state of the art method~\cite{A2ING}. This dataset is a subset of a real document stream from Yooz's customers. It is composed of 23 577 documents unequally distributed in 47 classes, 15 491 documents (65.71\%) of them are used for training, 2203 (9.34\%) for validation and 5883 (24,95\%) for test. Each class contains between 1 to 4075 documents. The distribution of documents between classes is illustrated by \textbf{Fig.\ref{fig2}}. The main drawback of this data-set is its unbalanced and incomplete test set. With it, it is impossible to avoid the possibility of statistical aberration, in particular for classes represented by only one or zero documents in the test set. In addition, classes with a weak representation does not really impact global performances even if its specified classification accuracy is null. In fact, only a small number of over-represented classes correctly classified is enough to get a high accuracy score.

\subsection{Evaluation method}
This section proposes to introduce the evaluation process apply in the next section. To evaluate each cases, we have to modify the training set consequently and only the training set. Classes impacted by the modification and documents used by the generated set was chosen randomly. For each run a new random set was generated according to the process described to the corresponding case. Obviously comparison results between methods are computed on the same randomly generated set.

Que les résultats de chaque méthodes sont envoyé à un système de rejet ce qui permet d'évaluer le niveau de définition de chaque classe par le système et à quel point ces définition lui permettent bien les différencier entre elle

In addition, an identical rejection system process each method results to evaluate the confidence level according to each class and the system capacity to distinguish classes between them even if the training is sub-optimal. The rejection system is an experimentally computed threshold applied on confidences scores obtained from the network. The last layer have one cell per class, and a sigmoid function provides the probability that the input image belong to each class. If the highest value is lower than the threshold, we reject the input in order to not miss-classify it. This is a deliberately simple rejection process because we wanted to evaluate only method performances and not the rejection system himself. 

As explained in introduction, a rejected document is considered as less important than a error. To include this in the evaluation, we choose to do not used classic F1-Score but a F0.5-score. It improves the importance of precision beside recall with a modification of \( \beta \) variable of F-score equation (with \( \beta = \) 0.5):

\begin{equation}
( 1 + \beta^2 ) \frac{Precision \cdot Recall}{ \beta^2 \cdot Precision + Recall }
\end{equation}

The other measures used are the Recall, the Precision and the Rejection Rate. The last measure is the global system accuracy that not includes the rejection system for its calculation, with the objective to display raw performances of neural network.

\section{Experiments}
\label{sec:experiments}
The proposed evaluation tends to compare different methods in four cases: ideal one with same number of document in training, validation and test ; unbalanced and incomplete cases where all classes are unbalanced or incomplete ; a realistic case which mimics a real document stream. This comparison aims to determine the efficiency and the adaptation of neural network for document stream classification. The first one is used as reference, the two next methods are corresponding to two specific stream difficulties, and the last one simulate a realistic document stream. The four next subsections are devoted to each of those cases, with a presentation of the case, of the modifications applied on the training dataset, an overall analysis of the results, a methods adaptation analysis and a table with results.

\subsection{Ideal case}
\label{sub-sec:Ideal}
In Ideal case, all the training set of RVL-CDIP is used for the training phase of each method. It corresponds to the traditional evaluation conditions of neural network, and we will use it as a benchmark for methods efficiency comparison.

Results display in \textbf{Table~\ref{IdealCaseTable}} show that the HCNN method got the best results. This can be explained by the fact that the dataset is more favorable to deep networks designed for image content recognition than the one dedicated to the textual content. Indeed, RVL-CDIP vocabulary is generally poor and many classes have a low number of words or in the worst case no words like the "file folders" class. In addition, an entire class and many other documents contain handwritten text that cannot be processed by the conventional OCR used here for the text extraction. Indeed, two classes are very complicated for text processing because they contain too few words and two others are also complicated but with slightly more words.

TCNN have the best performances for text content methods, it does not need too many words to work but the four complicated classes reduce drastically its accuracy. A low number of words is even more detrimental for RCNN that uses also as feature the word order inside the document text that need more text resources to be efficient. A2ING method suffer, in addition, of the huge number of documents because it was not designed to manage this. 

\begin{table}[ht]
\renewcommand{\arraystretch}{1.4}
\centering
\small
\begin{tabular}{|l|c|c|c|c|c|}
\hline
Value / Method & A2ING & HCNN & RCNN & TCNN \\
\hline
Accuracy & 31.06\% & \textbf{88.36\%} & 68.15\% & 79.67\% \\
\hline
Precision & \textbf{95.94\%} & 95.42\% & 86.10\% & 93.72\% \\
\hline
Recall & 23.53\% & \textbf{84.64\%} & 55.34\% & 69.91\% \\
\hline
F0.5-Score & 59.39\% & \textbf{93.05\%} & 77.48\% & 87.74\% \\
\hline
\end{tabular}
\caption{Result for Ideal case}
\label{IdealCaseTable}
\end{table}

\subsection{Unbalanced case}
\label{sub-sec:Unbalanced}
This second case simulates the unbalanced representation distribution between classes in a document stream. The objectives are to see if this unbalanced distribution impacts the neural network performances and how they are altered. The training set was modified with a reduction of some class distribution as follows: all classes are divided in four groups of four classes. Each group is linked to a percentage of their original number of documents and organize in tier. These tier are respectively tier 5\%, 10\%, 50\% and 100\%. This distribution is a modelization based on real document streams.

All systems are affected by the unbalanced training set and lose between 9\% and 11\% of their accuracy as introduced by \textbf{Table~\ref{UnbalancedCaseTable}}. The effect of this unbalanced training process highlights the over-fitting problem of the most represented classes (tier 100\%, 50\%) and an under-fitting (due to an insufficient training) for lesser represented classes (tier 10\%, 5\%). This unbalanced case results in a recall value higher than the precision for the first tier and the reverse for the second. The system seems to create trash classes with lesser well defined of tier 100\% classes. These classes have a very low precision (less than 50\%) in comparison of other (around 75\%). However, our proposed rejection system allows stabilizing the precision of high tier classes but to the detriment of low tier recall.

RCNN keep the worst result and it is the second in accuracy loss. It is the only method where the rejection system has entirely eliminate last tier classes (tier 5\%). The network has not been trained enough to manage these classes and their confidence scores were too low. HCNN is the most impacted method for accuracy reduction (-11.10\%) but it keeps the highest global performances. Last tier seems to be the biggest problem instead of tier 10\% that keep generally acceptable performances. TCNN is less impacted than RCNN and can handle the last tier unlike it.

\begin{table}[ht]
\renewcommand{\arraystretch}{1.4}
\centering
\small
\begin{tabular}{|l|c|c|c|c|c|}
\hline
Value / Method & HCNN & RCNN & TCNN \\
\hline
Accuracy & \textbf{78.26\%} & 57.08\% & 71.41\% \\
\hline
Precision & 89.14\% & 78.76\% & \textbf{91.63\%} \\
\hline
Recall & \textbf{76.25\%} & 51.68\% & 56.73\% \\
\hline
F0.5-Score & \textbf{86.22\%} & 71.29\% & 81.59\% \\
\hline
RRM Tier 5\% & 41.74\% & \textbf{84.85\%} & 55.98\% \\
\hline
RRM Tier 10\% & 30.31\% & \textbf{46.64\%} & 42.99\% \\
\hline
RRM Tier 50\% & \textbf{14.37\%} & 35.43\% & 58.53\% \\
\hline
RRM Tier 100\% & \textbf{8.60\%} & 26.35\% & 15.58\% \\
\hline
\end{tabular}
\caption{Result for Unbalanced case and Rejection Rate Means (RRM) for each classes tier}
\label{UnbalancedCaseTable}
\end{table}

\subsection{Incomplete case}
\label{sub-sec:Incomplete}
As explained in introduction of this article, it is impossible to generate a training set that contains all classes from a real case document stream. Any system used for document stream classification have to handle new/unexpected classes or at least reject them as noise. This test is probably the most difficult for neural network because they are absolutely not designed for this type of situation. The objectives is then to analyse in particular the rejection result for reduced classes, the effect of noises during training on complete class performances. For this case the training set is split in two equal groups. The first one corresponds to complete classes and uses all the documents for the training phase. The second group is the noisy classes. We considered those classes in a similar way to the one proposed in the one-shot learning challenge~\cite{OS1}\cite{OS2}\cite{NTM}\cite{SiamNet}. So only one document is used to train them.

As expected, the obtained results are bad in accordance to \textbf{Table~\ref{IncompleteCaseTable}}. No method has well classified even only one document from the noisy classes, and the rejection process did not balanced enough the impact of noise on complete class performances. Only some classes with a specific vocabulary and highly formatted structure keep high performances, the other become trash classes and gather all noisy classes documents. The rejection rate is higher for the noisy classes than the complete ones, with in average 46.35\% of documents rejected. But it is far from enough.

Again, the confidence scores of the RCNN approach do not allow differentiating enough the document classes, and the rejection system does not work well. The rejection rate is high for all classes and the noisy classes are not really more rejected than the others. Only one class obtained a good result as it is strongly defined by its vocabulary. TCNN and HCNN has got much better performances, in particular for specific vocabulary classes and for formatted structure classes. Moreover, they have an higher rejection rate for noisy classes.

\begin{table}[ht]
\renewcommand{\arraystretch}{1.4}
\centering
\small
\begin{tabular}{|l|c|c|c|c|c|}
\hline
Value / Method & HCNN & RCNN & TCNN \\
\hline
Accuracy & \textbf{45.99\%} & 40.86\% & 43.97\% \\
\hline
Precision & 60.75\% & 47.40\% & \textbf{61.38\%} \\
\hline
Recall & \textbf{71.66\%} & 68.89\% & 60.10\% \\
\hline
F0.5-Score & \textbf{62.66\%} & 50.55\% & 61.12\% \\
\hline
RRM Noise & 47.22\% & 36.03\% & \textbf{55.84\%} \\
\hline
RRM Complete & \textbf{9.46\%} & 26.19\% & 23.97\% \\
\hline
\end{tabular}
\caption{Result for Incomplete case and Rejection Rate Means (RRM) for each classes groups}
\label{IncompleteCaseTable}
\end{table}

\subsection{Realistic case}
\label{sub-sec:Realistic}
This scenario is a combination of the two previous cases, and was designed to be as close as possible to real documents stream. All classes are divided in five groups with in first time an incomplete group of four classes. In a second time, four groups of three classes with the same system than unbalanced case, with tier 5\%, 10\%, 50\% and 100\%. With this distribution we can finally simulate results of neural network methods on a document stream with an ideal test set and evaluate the global impact of document streams on neural network method efficiency.

We can observe in \textbf{Table~\ref{RealisticCaseTable}} that the obtained results are better than the ones from the incomplete case. This can be explained by the fact that fewer classes were incomplete. On the contrary, the obtained results are lower that the ones from the unbalanced case with the addition of noisy classes. There is no prominent differences between each unbalanced tier results and those from \textbf{\ref{sub-sec:Unbalanced}}, expect for the precision of 100\% tier classes that gather in addition noisy classes documents. On another hand, the rejection rate for incomplete classes is higher here, around 66.22\% on average. Like for \textbf{\ref{sub-sec:Incomplete}}, no documents of incomplete classes have been correctly classified. On the whole, all methods have lost between -23\% and -28\% of accuracy, so around one third of their original performances. Conclusions of individual result of each method is similar to the two previous cases.

The first tests with a artificial reduction of unbalanced seems to slightly improve performances of methods (with the rejection system). This modification result in a reduction of train samples for the 100\% tier, to equal the 50\% tier (so 6 classes with 50\% of their original training samples). This seems to support the importance of balancing classes in a neural network train set because at first glance, this reduction of training sample numbers should decrease the method performances.

\begin{table}[ht]
\renewcommand{\arraystretch}{1.4}
\centering
\small
\begin{tabular}{|l|c|c|c|c|c|}
\hline
Value / Method & A2ING & HCNN & RCNN & TCNN \\
\hline
Accuracy & 16.27\% & \textbf{62.46\%} & 44.70\% & 51.93\% \\
\hline
Precision & \textbf{84.63\%} & 77.55\% & 68.45\% & 78.50\% \\
\hline
Recall & 14.00\% & \textbf{71.82\%} & 41.98\% & 46.98\% \\
\hline
F0.5-Score & 42.13\% & \textbf{76.33\%} & 60.78\% & 69.21\% \\
\hline
RRM Noise & \textbf{97.20\%} & 55.49\% & 71.02\% & 72.16\% \\
\hline
RRM Tier 5\% & 87.85\% & 30.01\% & \textbf{89.29\%} & 66.79\% \\
\hline
RRM Tier 10\% & \textbf{85.20\%} & 29.13\% & 59.16\% & 63.87\% \\
\hline
RRM Tier 50\% & 77.81\% & \textbf{10.35\%} & 48.12\% & 33.08\% \\
\hline
RRM Tier 100\% & 78.19\% & \textbf{6.81\%} & 18.16\% & 22.83\% \\
\hline
\end{tabular}
\caption{Result for Realistic case and Rejection Rate Means (RRM) for each classes groups/tier}
\label{RealisticCaseTable}
\end{table}

\section{Conclusion}
\label{sec:conclusion}

\subsection{Results Summary}
In a general analysis based on \textbf{Table~\ref{SummaryTableRVL-CDIP}}, we can say that neural network classification systems are unreliable in the situation of industrial document stream. They cannot handle very low represented or unexpected classes. They can deal, with difficulty, slightly more represented classes and they have high result for over-represented classes even if the precision was reduced by sub-represented classes that they gathered. The impact of incomplete classes is the main problem with the impossibility to add new unexpected class encountered to the classification system. This seems to be unsolvable without an important structural adaptation. 

The unbalanced number of document per class inside the training set reduces the performances. The neural network based systems then tends to over-fit their model for the highest represented classes, to the detriment of the lesser well defined ones. The less represented classes are affected by sub-optimal training. In another hand, the unbalanced distribution of classes problem can be subdued by an adaptation of the training set. For instance, reducing the gap between the highest represented classes and the lowest represented classes, has a positive effect on the thresholded performances (\emph{i.e.} the ones obtained with the rejection process) by reducing the over-fitting problem and getting a better modelization between them. The next step for this ways could be to improve the representation of low classes by the generation of samples like in \cite{OS1}. 

\begin{table}[ht]
\renewcommand{\arraystretch}{1.4}
\centering
\small
\begin{tabular}{|l|c|c|c|c|c|c|c|c|c|}
\hline
RVL-CDIP & \multicolumn{3}{c|}{A2ING} & \multicolumn{3}{c|}{HCNN} \\
\hline
Case / Method & Acc & Pre & Rec & Acc & Pre & Rec \\
\hline
Ideal & 31.06\% & \textbf{95.94\%} & 23.53\% & \textbf{88.36\%} & 95.42\% & \textbf{84.64\%} \\
\hline
Unbalanced & --- & --- & --- & \textbf{78.26\%} & 89.14\% & \textbf{76.25\%} \\
\hline
Incomplete & --- & --- & --- & \textbf{45.99\%} & 60.75\% & \textbf{71.66\%} \\
\hline
Realistic & 16.27\% & \textbf{84.63\%} & 14.00\% & \textbf{62.46\%} & 77.55\% & \textbf{71.82\%} \\
\hline
\hline
RVL-CDIP & \multicolumn{3}{c|}{RCNN} & \multicolumn{3}{c|}{TCNN } \\
\hline
Case / Method & Acc & Pre & Rec & Acc & Pre & Rec \\
\hline
Ideal & 68.15\% & 86.10\% & 55.34\% & 79.67\% & 93.72\% & 69.91\% \\
\hline
Unbalanced & 57.08\% & 78.76\% & 51.68\% & 71.41\% & \textbf{91.63\%} & 56.73\% \\
\hline
Incomplete & 40.86\% & 47.40\% & 68.89\% & 43.97\% & \textbf{61.38\%} & 60.10\% \\
\hline
Realistic & 44.70\% & 68.45\% & 41.98\% & 51.93\% & 78.50\% & 46.98\% \\
\hline
\end{tabular}
\caption{Summary Table RVL-CDIP}
\label{SummaryTableRVL-CDIP}
\end{table}

\begin{table}[ht]
\renewcommand{\arraystretch}{1.4}
\centering
\small
\begin{tabular}{|l|c|c|c|c|c|c|c|c|c|}
\hline
Private D-S & \multicolumn{3}{c|}{A2ING} & \multicolumn{3}{c|}{HCNN} \\
\hline
Case / Method & Acc & Pre & Rec & Acc & Pre & Rec \\
\hline
True & 85.37\% & 97.88\% & 80.79\% & 89.26\% & 98.13\% & 88.39\% \\
\hline
\hline
Private D-S & \multicolumn{3}{c|}{RCNN} & \multicolumn{3}{c|}{TCNN } \\
\hline
Case / Method & Acc & Pre & Rec & Acc & Pre & Rec \\
\hline
True & 88.63\% & 95.67\% & 78.83\% & \textbf{93.44\%} & \textbf{98.97\%} & \textbf{89.28\%} \\
\hline
\end{tabular}
\caption{Summary Table Private Data-set}
\label{SummaryTablePrivateDS}
\end{table}

\subsection{Neural Networks Robustness Conclusion}
On the whole, neural network classification systems are highly preforming in case where training data was not a problem. But for document stream classification, they cannot manage low represented and unexpected classes with same performances. Although, this type of classes represents more than a half of all stream classes. 

Results introduce by \textbf{Table~\ref{SummaryTablePrivateDS}} computed on the private Yooz's dataset do not lead to the same conclusion, all methods performances are high because the test set composition hide low classes impact on global result. Indeed, if the evaluation set is incomplete and unbalanced like the training set, low represented classes are impossible to evaluate, with sometimes only one are two documents. Individual class performance are not reliable for this dataset because low classes have not enough test samples. In addition, this too few samples does not impact global performances. In fact, its is possible to achieved a 70\% accuracy with only the classification of three main classes (\textbf{Fig.\ref{fig2}}).

In an other hand, this dataset was build for the evaluation of text approach method and HCNN offer a surprisingly high score on it despite is composition mostly verbose. The RVL-CDIP dataset are probably not the best choice to compare text and image approaches, it was clearly designed for image classification method.

\subsection{Perspectives}
We need more test to determine the limit of training samples required by each method to stay reliable. Equally, it can be interesting to determine a formula that describe the behaviour of network efficiency as function of unbalance rate. One-shot learning methods like NTM and MANN \cite{NTM} could be a option to balance impact of incomplete or very low represented classes on performances. The best solution seems to be the apart classification of this type of classes. 

Finally, text or image approaches are both highly performing but not on same classes and the weakness of one could be balanced by the strength of the other on this dataset. A good multi-modal system could give better result as same as an improvement of rejection system.


\bibliographystyle{abbrv}
\bibliography{biblio}
\end{document}